%% file: main.tex
\documentclass[10pt,twocolumn,letterpaper]{article}

\usepackage[pagenumbers]{paperstyle} 

\input{preamble}

\usepackage{bbm}
\usepackage[table]{xcolor}
\definecolor{linkblue}{rgb}{0.21,0.49,0.74}
\usepackage[pagebackref,breaklinks,colorlinks,allcolors=linkblue]{hyperref}

\def\psPaperID{} 
\def\confName{}
\def\confYear{}

\title{MEDit-Bench: A Dataset for Evaluating Message-Driven Narrative Video Editing}

\author{%
Katsuya Ogata$^{1}$ \quad Zongshang Pang$^{1}$ \quad Mayu Otani$^{2}$ \quad Yuta Nakashima$^{1}$\\
$^{1}$The University of Osaka\\
$^{2}$CyberAgent, Inc.\\
{\tt\small ogata@im.sanken.osaka-u.ac.jp}\,\,{\tt\small pangzs@im.sanken.osaka-u.ac.jp}\\
{\tt\small otani\_mayu@cyberagent.co.jp}\,\,{\tt\small n-yuta@im.sanken.osaka-u.ac.jp}%
}

\begin{document}
\maketitle

\input{sec/0_abstract}
\input{sec/1_intro}
\input{sec/2_related}

\input{sec/2_dataset}

\input{sec/3_method}
\input{sec/4_experiments}

\input{sec/7_limitations}
\input{sec/5_conclusion}
{
    \small
    \bibliographystyle{ieeenat_fullname}
    \bibliography{main}
}

\clearpage
\maketitlesupplementary
\input{sec/8_supplementary}

\end{document}

%% file: preamble.tex
\usepackage{graphicx}
\usepackage{amsmath}
\usepackage{amssymb}
\usepackage{subcaption}
\usepackage{booktabs}
\usepackage{threeparttable}
\usepackage{multirow}
\usepackage{tikz}
\usepackage{stfloats}

\newcommand{\winlossbar}[2]{%
	$\vcenter{\hbox{%
	\begin{tikzpicture}[baseline=0pt]
		\pgfmathsetmacro{\bw}{1.7}
		\pgfmathsetmacro{\bh}{0.32}
		\pgfmathsetmacro{\ww}{#1/100*\bw}
		\fill[green!30] (0,0) rectangle (\ww,\bh);
		\fill[red!30] (\ww,0) rectangle (\bw,\bh);
		\node[inner sep=0,anchor=center] at (\bw/2,\bh/2)
			{\scriptsize\textbf{#1\%}};
		\draw[black!50] (0,0) rectangle (\bw,\bh);
	\end{tikzpicture}%
	}}$%
}

\usepackage{listings}
\usepackage{xcolor}
\usepackage{tcolorbox}
\tcbuselibrary{listings,skins,breakable}

\definecolor{prompthead}{RGB}{210,228,245}
\definecolor{promptvar}{RGB}{180,60,0}

\newtcblisting{promptbox}[1]{
  listing only,
  listing options={
    basicstyle=\ttfamily\scriptsize,
    breaklines=true,
    breakatwhitespace=false,
    columns=flexible,
    keepspaces=true,
    showstringspaces=false,
    breakindent=0pt,
  },
  enhanced,
  breakable,
  colback=gray!4,
  colframe=gray!25,
  colbacktitle=prompthead,
  coltitle=black,
  fonttitle=\bfseries\normalsize\centering,
  title={#1},
  arc=5pt,
  boxrule=0.5pt,
  titlerule=0.3pt,
  top=2pt, bottom=2pt,
}

%% file: sec/0_abstract.tex
\begin{abstract}
Video editing is fundamentally message-driven: even from the same source footage, the selected shots change depending on the narrative the editor wishes to convey.
Benchmarks for a closely related task, video summarization, reduce editorial intent to a single, message-agnostic notion of saliency and thus do not account for this diversity.
For evaluating message-driven video editing, we present \textbf{MEDit-Bench}, a dataset and benchmark, which 
pairs long-form videos with multiple editing messages and multiple professionally produced edits per message, demonstrating that different messages yield substantially different edits from the same source.
We define an automatic evaluation protocol based on temporal alignment metrics, and find that an LLM-as-a-judge preference, a natural proxy for narrative quality, is unreliable for this task due to severe position bias.
We additionally annotate each message with ambiguity and contextfulness scores, and show that both dimensions negatively correlate with model performance, establishing message difficulty as a meaningful stratification factor.
Experiments with state-of-the-art MLLMs and reinforcement fine-tuned baselines show that while strong models approach human temporal alignment at lenient thresholds, all models fall behind humans at stricter criteria.
A human perceptual study further confirms a large quality gap, with professional human edits remaining consistently preferred over model outputs.
The project page is available at \url{https://ogatakatsuya.github.io/medit-bench}.

\end{abstract}

%% file: sec/1_intro.tex
\section{Introduction}
\label{sec:intro}

\begin{figure}[htbp]
  \centering
  \includegraphics[width=\columnwidth]{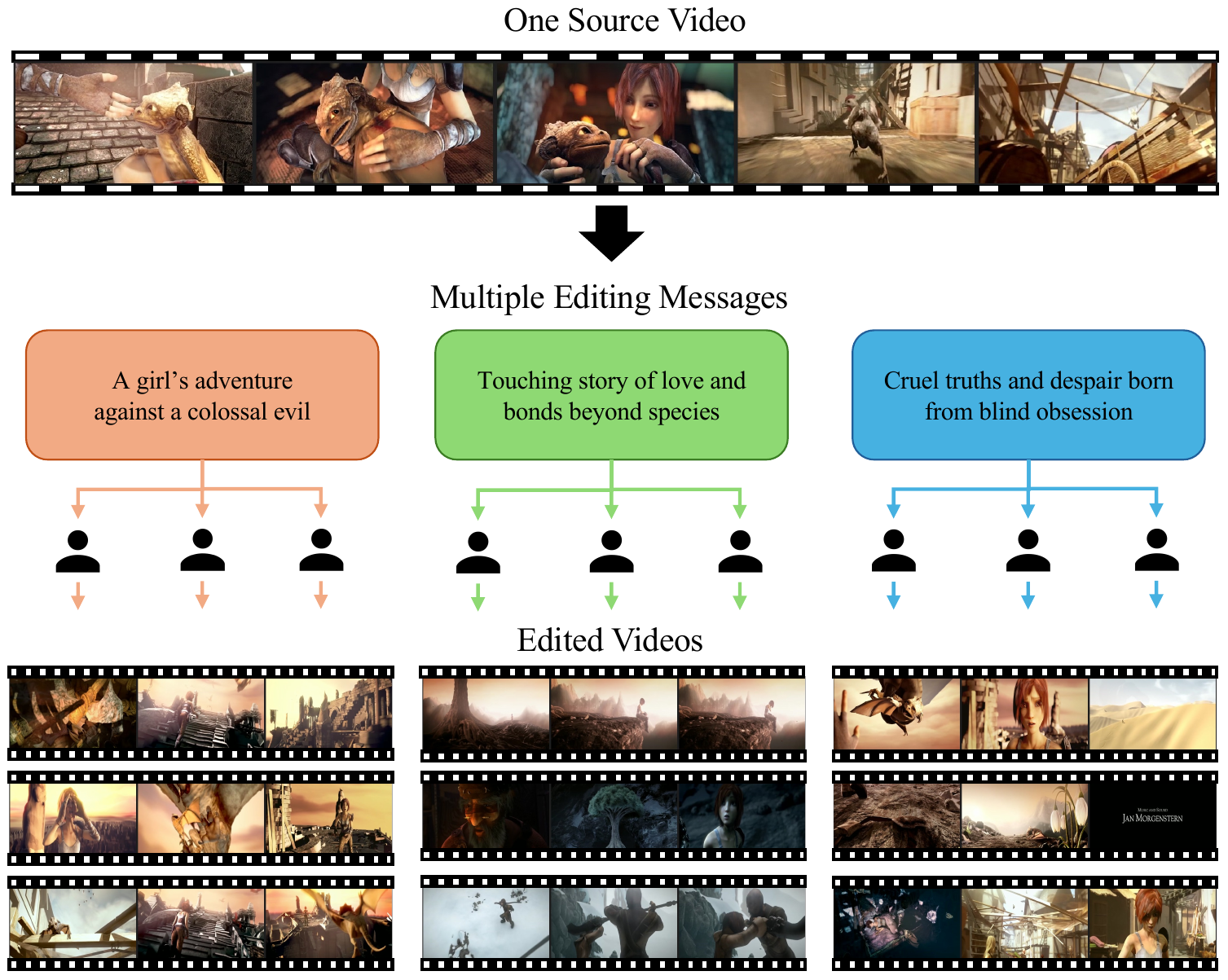}
  \caption{\textbf{Overview of MEDit-Bench.}
    Given a long-form source video, multiple editing messages specify different narratives.
    For each (video, message) pair, multiple professional editors independently produce a cut-only edit.
    The same source footage results in different cut sequences depending on the specified message, highlighting the diversity that existing summarization benchmarks do not capture.
  }
  \label{fig:teaser-single}
\end{figure}

Online videos have become a primary medium for information, education, and entertainment.
As a result, there is a growing demand for automating video editing, which transforms raw footage into watchable content.
Video summarization has been extensively studied as a step toward automation, supported by datasets such as SumMe~\cite{10.1007/978-3-319-10584-0_33}, TVSum~\cite{7299154}, and VSUMM~\cite{DEAVILA201156}, tailored to evaluate models' ability to find salient segments. 
Representative methods include supervised importance prediction~\cite{10.1145/3394171.3414064,9666088,10657498}, vision-language approaches leveraging CLIP-like models~\cite{NEURIPS2021_7503cfac,he2023a2summ,li2023progressive,10031001}, and LLM-based frameworks~\cite{lee2025videosummarizationlargelanguage}.

However, video editing is not merely about selecting salient segments.
Unlike saliency-based summarization, where the goal is to identify objectively important moments, professional editing is inherently purposeful: it requires deep comprehension of long-form video content and the creative judgment to compose shots into a sequence that tells a story.
The resulting edit is therefore not a compression of the footage but a \emph{deliberate creative act}: it demands both an understanding of what is happening in the video and a deliberate choice of how to present it to communicate something meaningful to viewers.

The key ingredient that drives this process is the \emph{message}, a natural language statement of the narrative the editor wishes to convey.
The message determines not just \emph{which} moments to include but how they should be composed into a coherent story: a visually salient moment may be entirely irrelevant to a given message, while an apparently unremarkable shot may be indispensable for the narrative setup.
Crucially, the same source footage can yield radically different edits under different messages, and even for the same message, multiple valid interpretations exist that reflect individual editorial style.
This diversity (across messages and across editors) is absent from existing video summarization benchmarks, which primarily focus on summaries of visual highlights~\cite{10.1007/978-3-319-10584-0_33, 7299154} and thus conflate the richness of editorial intent into a single notion of saliency.

In this work, we focus on \emph{message-driven video editing} and make the following contributions.
\textbf{(1)} We formulate the task and introduce \textbf{MEDit-Bench} (\textbf{ME}ssage-\textbf{D}riven video ed\textbf{it}ing \textbf{Bench}mark), which pairs 60 long-form videos with 540 professional message-conditioned edits, providing multiple messages per video and multiple editors per message (\cref{fig:teaser-single}) to capture the diversity of editorial intent that single-reference summarization benchmarks overlook.
\textbf{(2)} We establish an evaluation protocol centered on temporal alignment metrics (R@$\theta$, F1@$\theta$, mIoU), stratified by message quality (ambiguity and contextfulness) and complemented by a 1{,}620-judgment human perceptual study; we further show that an LLM-as-a-judge preference metric~\cite{10.5555/3666122.3668142,gu2024survey,zhang2023gpt4visiongeneralistevaluatorvisionlanguage} suffers from severe position bias and is unreliable for this task, a cautionary result for automatic narrative-quality evaluation.
\textbf{(3)} We propose \textbf{MEDitor}, a model trained with GRPO using editing-specific rewards (temporal coverage and cut-level F1); MEDitor-8B surpasses all zero-shot open-source baselines and rivals commercial models.
\textbf{(4)} Through extensive experiments, we find that even strong MLLMs (Gemini-3-pro, GPT-5, Qwen3-VL) fall well short of the human ceiling at strict cut-precision thresholds, a gap corroborated by our human perceptual study in which professional edits remain consistently preferred over model outputs, while MEDitor's editing ability transfers zero-shot to classic video summarization (SumMe, TVSum), matching or exceeding prior unsupervised methods.

%% file: sec/2_related.tex
\section{Related Work}
\label{sec:related}

\subsection{Video Editing and Summarization Datasets}
Video editing and summarization datasets have been developed for a variety of objectives and modalities.
Early video summarization benchmarks, SumMe~\cite{10.1007/978-3-319-10584-0_33}, TVSum~\cite{7299154}, and VSUMM~\cite{DEAVILA201156}, score clip-level importance based on visual saliency and user interest, and are widely used for evaluating importance-based segment selection.
More recently, multimodal datasets have emerged that pair video with text descriptions for semantic summarization~\cite{lin2023videoxum, Qiu2023MMSumAD}, broadening the scope of textual signals in evaluation.

Datasets focusing on movie analysis and editing technique~\cite{huang2020movie, pardo2022moviecuts, 9710256} have advanced shot boundary detection and structural analysis of professional film editing.
However, these datasets address individual editing components in isolation and do not model the end-to-end editing process conditioned on a communicative message.
In contrast, MEDit-Bench is centered on the message that an editor wishes to convey.
Single source video is paired with multiple natural-language messages, and professional editors independently produce edits for each, enabling evaluation of message-conditioned diversity, which is a dimension absent from existing benchmarks.


\subsection{Video Summarization and Editing Methods}
Video summarization has been the predominant target of automated video editing.
Early work used CNNs and RNNs to assign importance scores to frames or shots~\cite{fajtl2018summarizing, zhu2020dsnet, 10.1145/3394171.3414064, 9666088, 9428318, 9879839, 10657498}.
More recent methods leverage vision-language models, such as CLIP, to incorporate textual cues and improve multimodal alignment~\cite{NEURIPS2021_7503cfac, he2023a2summ, li2023progressive, 10031001}.
An LLM's chain-of-thought reasoning has been used for frame-wise importance scoring, further pushing the state of the art~\cite{lee2025videosummarizationlargelanguage}.
Although these methods achieve strong performance on importance-based benchmarks, they optimize for a fixed notion of ``important'' content and do not support message-conditioned narrative construction.

Video editing has been studied from different perspectives.
Methods for aligning narration or dialogue scripts with footage temporally~\cite{10.1145/2984511.2984569, 10.1145/3072959.3073653}, text-based clip retrieval for montage generation~\cite{Wang2019writeavideo, xiong2023transcriptvideoefficientclip}, and systems that transfer professional editing styles or generate full timelines from natural-language instructions~\cite{frey2021automaticnonlinearvideoediting, pardo2024generativetimelinesinstructedvisual} each address specific aspects of the editing pipeline. 
Unlike these existing works, our task requires the model to holistically interpret a long video and autonomously construct a narrative aligned with the given message.




\subsection{Video Temporal Localization with MLLMs} Video temporal localization involves a suite of related tasks such as temporal grounding \cite{lei2021detecting}, action localization and video summarization \citep{10.1007/978-3-319-10584-0_33, 7299154, pang2023contrastive, DRDSN}, which invariably require the model to find the segments that best correspond to the input query. Recently, MLLMs have been leveraged to tackle such tasks, with most of them focusing on developing MLLM-friendly representations of time, including time-aware positional embedding \cite{ren2024timechat, guo2025vtg}, semantic-based embedding matching \cite{liu2024bench, pang2026measure}, and interleaved textual timestamps \cite{meinardus2024chrono}, where models are mostly trained with supervised fine-tuning. As the time representation starts to converge to the interleaved timestamps \cite{Qwen3-VL, zhang2026timelens}, recent works have begun to study the optimization strategy, \ie, the shift from supervised fine-tuning to reinforcement learning \cite{wang2026time, zhang2026timelens}. 

Message-driven video editing requires the same temporal localization ability but is harder in two ways. First, rather than retrieving segments that explicitly match the query, it must infer those that are contextually relevant to the message, \ie, that set up, explain, or advance it, not merely depict its surface content. Second, while existing multi-segment tasks output segments that are either near-duplicates of one semantic concept (\eg, action localization) or mutually independent (\eg, dense captioning \cite{krishna2017dense, yang2023vid2seq}, video summarization), ours are contextually connected into a coherent storyline. For instance, an edit for ``The revenge of the white rabbit'' may include segments establishing why the rabbit seeks revenge, not only the revenge scenes.

%% file: sec/2_dataset.tex
\section{MEDit-Bench}
\label{sec:dataset}

We design MEDit-Bench to evaluate a model's ability to generate video narratives from a high-level message, which requires deep comprehension of the video content and the creation of a convincing storyline that conveys the message. 

\subsection{Data Collection}
MEDit-Bench is constructed in three stages: (i)~source video selection, (ii)~editing message creation, and (iii)~professional editing and annotation.

\noindent\textbf{Source videos and messages.}
We select 60 videos of length 7--18 minutes from LongVideoBench~\cite{wu2024longvideobenchbenchmarklongcontextinterleaved}, VideoMME~\cite{fu2025video}, and Blender Studio open movies~\cite{blender_studio_films}.
For each source video, we write three distinct editing messages that describe the narrative the editor wishes to convey to viewers, resulting in 180 video-message pairs.
Messages are designed to be (i)~visually communicable without relying on audio or dialogue, (ii)~diverse in intent so that they encourage substantially different edits from the same footage, and (iii)~varied in difficulty, spanning clear and concrete messages to ambiguous or context-dependent ones.

\noindent\textbf{Professional annotation.}
During the annotation process, each video-message pair is given to three professional video editors, who independently produce a collection of temporal cuts, each of which is represented by the start and end timestamps of a selected clip, resulting in $540$ edited videos in total; further details on dataset construction, statistics, and representative examples are provided in the supplementary material.
Editors are instructed to (i) trim the source to a 1--3 minute summary using only cut operations, (ii) reorder shots freely, (iii) ignore the audio and base their decisions solely on the visual content, and (iv) provide a rationale for every cut to discourage careless cuts. Although the shot ordering and rationales are not used in our benchmark, we release them to the community for future research.


\subsection{Comparison with Existing Datasets}
\Cref{tab:comparison} compares MEDit-Bench with existing summarization datasets.
Unlike prior datasets that provide a single target edit per source video without an explicit message, MEDit-Bench provides multiple messages and multiple edits per message, enabling evaluation of the diversity stemming from different messages as well as editors' styles and tastes.
MEDit-Bench also features substantially longer source footage (13m44s on average) compared to SumMe (2m40s), VSUMM (1m30s), and TVSum (4m12s), posing a greater challenge for models to process.

\begin{table}[tb]
  \centering
  \caption{Comparison with representative video summarization datasets.}
  \label{tab:comparison}
  \resizebox{\columnwidth}{!}{%
  \begin{tabular}{lcccc}
    \toprule
    \textbf{Dataset} & \textbf{Message} & \textbf{Footage} & \textbf{Edits} & \textbf{Avg. duration} \\
    \midrule
    SumMe~\cite{10.1007/978-3-319-10584-0_33} & $\times$  & 25 & 25 & 2m40s  \\
    VSUMM~\cite{DEAVILA201156}                 & $\times$  & 50 & 50 & 1m30s  \\
    TVSum~\cite{7299154}                       & $\times$ & 50 & 50 &  4m12s  \\
    \midrule
    \textbf{MEDit-Bench} & $\checkmark$ & \textbf{60} & \textbf{540} & \textbf{13m44s} \\
    \bottomrule
  \end{tabular}}
\end{table}

We analyze agreement across editors (cross-annotator) and across messages (cross-message), following the existing protocol~\cite{8954229}.
\Cref{tab:f1-comparison} shows that cross-annotator F1-Avg is 0.42 (higher than SumMe at 0.31 and lower than TVSum at 0.54), reflecting editors' different styles.
Cross-message F1-Avg is only 0.15, indicating that messages substantially diversify the edits even for the same editor.

\begin{table}[tb]
  \centering
  \caption{Agreement across editors (Cross-Ann.) and across messages (Cross-Msg.) measured by F1 score.
  $^1$Cross-Ann.: agreement between different editors for the same video and message.
  $^2$Cross-Msg.: agreement between different messages for the same video and editor.
  $^*$Quoted from~\cite{8954229}.}
  \label{tab:f1-comparison}
  \resizebox{\columnwidth}{!}{
  \begin{tabular}{cccccccc}
    \toprule
    \multicolumn{2}{c}{\textbf{SumMe}~\cite{10.1007/978-3-319-10584-0_33}} & \multicolumn{2}{c}{\textbf{TVSum}~\cite{7299154}} & \multicolumn{4}{c}{\textbf{MEDit-Bench (ours)}} \\
    \cmidrule(lr){1-2} \cmidrule(lr){3-4} \cmidrule(lr){5-8}
    & & & & \multicolumn{2}{c}{Cross-Ann.$^1$} & \multicolumn{2}{c}{Cross-Msg.$^2$} \\
    \cmidrule(lr){5-6} \cmidrule(lr){7-8}
    Avg$^*$ & Max$^*$ & Avg$^*$ & Max$^*$ & Avg & Max & Avg & Max \\
    \midrule
    0.31 & 0.54 & 0.54 & 0.78 & 0.42 & 0.55 & \textbf{0.15} & \textbf{0.27} \\
    \bottomrule
  \end{tabular}
  }
\end{table}

\subsection{Message Quality Annotations}
\label{sec:msg_labels}

To characterize the inherent characteristics of messages, we annotate each of the 180 messages in the following two aspects using an LLM-assisted protocol:

\noindent\textbf{Ambiguity} (1--5, low to high): To what extent does the message admit multiple valid interpretations?
A highly ambiguous message admits multiple equally valid narrative interpretations, making it harder for any single edit to match the reference (\eg, \textit{``A story of resilience''} provides no concrete visual anchor and can be realized in many ways).

\noindent\textbf{Contextfulness} (1--5, low to high): How much context is required to understand the message?
A contextful message can require some introductory cuts to make the narrative meaningful (\eg, \textit{``The long-awaited reunion''} presupposes that the viewer understands who separated and why, requiring contextual buildup before the key moment).

\begin{figure}[tb]
  \centering
  \includegraphics[width=\linewidth]{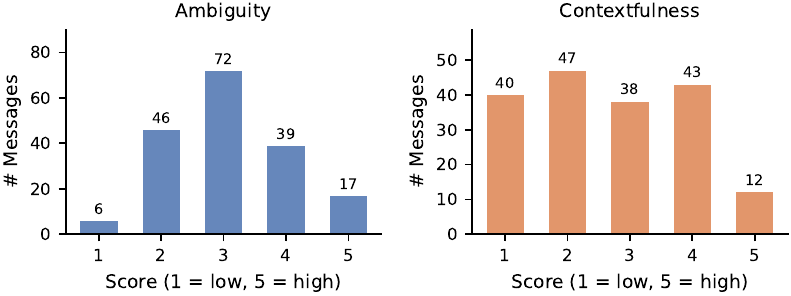}
  \caption{Distribution of message quality annotations across the 180 MEDit-Bench messages.
  Ambiguity measures how open the message is to interpretation.
  Contextfulness measures the extent to which external context is required.}
  \label{fig:label_dist}
\end{figure}

\Cref{fig:label_dist} shows the distribution of both properties across the 180 messages.
Most messages fall in the moderate-to-high range for ambiguity (scores 3--4) and in the low-to-moderate range for contextfulness (scores 1--2), indicating that our messages are generally interpretable but leave some room for creative variation.

\subsection{Task Definition}
  Given a video $V$ and a message $M$ in natural language text, we propose to predict the edited video as a sequence of temporal cuts
  \begin{align}
    \hat{\mathbf{C}} = (c_1, c_2, \ldots, c_K),
  \end{align}
  where $c_k = (s_k,\, e_k)$, with $s_k$ and $e_k$ ($0 \le s_k < e_k \le T$, with $T$ the video duration) denoting the start and end timestamps of the $k$-th cut.
  Unlike conventional summarization, which targets a single general synopsis per video, our task is explicitly conditioned on $M$, which is arbitrarily specified by the user, enabling $\hat{\mathbf{C}}$ to convey diverse narratives. 

\subsection{Evaluation Metrics}
\label{sec:evaluation}

We propose to evaluate predicted edit $\hat{\mathbf{C}}$ on the MEDit-Bench's triplet $\mathcal{D} = \{(V, M, \mathbf{C})\}$ using the following three metrics, where $V$, $M$, and $\mathbf{C} = (c_1, \ldots, c_{K'})$ represent the footage, message, and reference cuts, respectively. 

\noindent\textbf{Recall at Temporal IoU.} 
Since each edit consists of multiple cuts, we adopt a \emph{union-based} temporal IoU formulation that treats a cut sequence as a single temporal region. Let $\mathrm{tIoU}(c,c')=|c\cap c'|/|c\cup c'|$ be the temporal IoU of two cuts, where $|\cdot|$ denotes the duration,
we define a binary hit at threshold $\theta$ as
\begin{align}
  h_\theta(\hat{\mathbf{C}},\, \mathbf{C}) = \mathbbm{1}\!\left[\text{tIoU}(\hat{\mathbf{C}},\, \mathbf{C}) \ge \theta\right],
\end{align}
where $\text{tIoU}(\hat{\mathbf{C}},\, \mathbf{C}) = \mathrm{tIoU}\!\Bigl(\textstyle\bigcup_m \hat{c}_m,\ \bigcup_n c_n\Bigr)$. R@$\theta$ is the hit rate averaged over $\mathcal{D}$:
\begin{align}
  \text{R@}\theta = \frac{1}{|\mathcal{D}|} \sum_{\mathcal{D}} h_{\theta}(\hat{\mathbf{C}},\, \mathbf{C}),
\end{align}
which we report at $\theta \in \{0.3, 0.5, 0.7\}$ along with the mean temporal IoU (mIoU).

\noindent\textbf{F1 Score with Greedy Cut Matching.}
We collect all candidate pairs $(i, j)$ with $\text{IoU}(\hat{c}_i, c_j) \geq \theta$, sort them by IoU in descending order, and assign each predicted and reference cut to at most one partner via greedy matching, yielding a matched set $\mathcal{M}_\theta \subseteq [K] \times [K']$ ($[N]$ denotes the set of positive integers up to $N$). F1 at threshold $\theta$ are defined as
\begin{align}
 \text{F1}@\theta = \frac{2\,P@\theta\cdot R@\theta}{P@\theta + R@\theta},
\end{align}
where $\mathrm{P@\theta}=|\mathcal{M}_\theta|/K$ and $\mathrm{R@\theta}=|\mathcal{M}_\theta|/K'$ and $\theta \in \{0.3, 0.5, 0.7\}$. Unlike the union-based recall above, the F1 score also evaluates the per-cut alignment and cut quantity consistency between the predicted and the reference sequence.

\noindent\textbf{LLM-based Preference.}
Temporal metrics alone are insufficient: an edit $\hat{\mathbf{C}}$ may convey the message well without temporally matching $\mathbf{C}$, or match it closely yet miss key context~\cite{liu-etal-2023-g}.
As an experimental probe of \emph{semantic} quality, we adopt an LLM-as-a-judge protocol~\cite{10.5555/3666122.3668142,gu2024survey} with Gemini-3-pro.
The judge performs a blind A/B comparison of $\hat{\mathbf{C}}$ and $\mathbf{C}$ given $M$, considering shot selection and storytelling flow.
Let $\mathcal{J}_M(\mathbf{C}, \hat{\mathbf{C}}) \in \{0, 1\}$ return $1$ if the first-presented edit is preferred.
Since LLM judges exhibit \emph{position bias}~\cite{position-bias}, we evaluate each pair in both orders and count a win only when $\hat{\mathbf{C}}$ is preferred regardless of position:
\begin{align}
    \text{Agg}_M(\hat{\mathbf{C}}, \mathbf{C}) = \mathcal{J}_M(\hat{\mathbf{C}}, \mathbf{C}) \times (1-\mathcal{J}_M(\mathbf{C}, \hat{\mathbf{C}})).
\end{align}
The WinRate is the mean over the dataset:
\begin{align}
  \text{WinRate} = \frac{1}{|\mathcal{D}|} \sum_{(V, M, \mathbf{C}) \in \mathcal{D}} \text{Agg}_M(\hat{\mathbf{C}}, \mathbf{C}).
\end{align}

%% file: sec/3_method.tex
\section{Methods}
To tackle the proposed message-driven video editing task, we propose a training-free baseline method using frozen MLLMs  (\cref{method: frozen}) and a reinforcement learning method trained on pseudo-labels distilled from a frontier MLLM (\cref{method: grpo}).

\subsection{Editing with Frozen MLLMs}
\label{method: frozen}

MLLMs such as LLaVA \cite{liu2023visual} and the Qwen-VL family \cite{Qwen3-VL} have been shown to excel in video tasks such as video QA and captioning. Recent works have also revealed their potential in video temporal localization tasks such as
temporal grounding and video summarization \cite{ren2024timechat, pang2026measure, liu2024bench}. To probe the message-driven video editing capability of such MLLMs, we treat an MLLM as a frozen function $f$
that directly maps the video and message to a sequence of cuts:
\begin{align}
\hat{\mathbf{C}} = f(V, M).
\end{align}
The MLLM is prompted with a fixed instruction (see the supplementary material for the full prompt and configuration) that introduces the task and specifies the output format.

\subsection{GRPO with Editing Rewards}
\label{method: grpo}

Group relative policy optimization (GRPO)~\cite{deepseek-math} is widely used to train MLLMs on video tasks~\cite{feng2026video, li2025videochat, wang2026time, zhang2026timelens}; for temporal grounding, Time-R1~\cite{wang2026time} and TimeLens~\cite{zhang2026timelens} reward the temporal intersection-over-union (tIoU) between the predicted and reference span.
We adopt the same cut format but find a tIoU-only reward suboptimal for editing: grounding localizes a \emph{single} reference span, whereas an edit is a \emph{sequence} of cuts.
Extending tIoU to the union of the predicted and reference cuts rewards only \emph{which} portion of the timeline is selected, not \emph{how} it is partitioned, leaving the number and boundaries of cuts unsupervised; a few coarse cuts can then match the reference's coverage but not its composition.
We therefore complement this coverage term with a \emph{cut-level F1} reward that matches predicted and reference cuts one-to-one, rewarding the correct number and placement of cuts.

\paragraph{Reward.}
Following \cref{sec:evaluation}, let $\hat{\mathbf{C}}=(\hat{c}_1,\dots,\hat{c}_{K})$ and $\mathbf{C}=(c_1,\dots,c_{K'})$ be the predicted and reference cuts of a training triplet $(V, M, \mathbf{C}) \in \mathcal{D}_\text{GRPO}$.
Our \emph{coverage} reward is the temporal IoU between the unions of the two cut sets,
\begin{equation}
\mathcal{R}_{\mathrm{cov}} = \text{tIoU}(\hat{\mathbf{C}}, \mathbf{C}), \label{eq:tiou_reward}
\end{equation}
which scores temporal overlap but is invariant to how the union is partitioned into cuts.

The \emph{cut-level F1} reward instead operates on individual cuts: we form a greedy matching set $\mathcal{P}\subseteq [K] \times [K']$ from the prediction and reference cut pairs like described in~\cref{sec:evaluation} but without the tIoU-based thresholding, and then accumulate soft true positives $\text{TP}=\sum_{(m,n)\in\mathcal{P}}\mathrm{tIoU}(\hat{c}_m,c_n)$, yielding 
\begin{equation}
\mathcal{R}_{\mathrm{F1}}=\frac{2\,\mathrm{P}\cdot\mathrm{R}}{\mathrm{P}+\mathrm{R}},
\end{equation}
where $\mathrm{P}=\text{TP}/K$ and $\mathrm{R}=\text{TP}/K'$.
This soft, tIoU-weighted $\text{TP}$, rather than a thresholded count, keeps the reward continuous in the cut boundaries.

The final reward combines these two terms with equal weight, \ie,
\begin{equation}
\mathcal{R}=\mathcal{R}_{\mathrm{cov}}+\mathcal{R}_{\mathrm{F1}},
\end{equation}
so that $\mathcal{R}_{\mathrm{cov}}$ enforces a decent global coverage while $\mathcal{R}_{\mathrm{F1}}$ the correct number and placement of cuts. We ablate the effect of these two rewards in the supplementary material.

\paragraph{Policy Optimization.}
We optimize the policy $\pi_\Phi$ with GRPO~\cite{deepseek-math}, which directly maximizes the reward $\mathcal{R}$ rather than imitating the reference cuts by maximum likelihood.
For each training sample $(V, M, \mathbf{C}) \in \mathcal{D}_\text{GRPO}$, we sample a group $\mathcal{G} = \{\hat{\mathbf{C}}\}$ of $|\mathcal{G}|$ candidate edits from the current policy and score each with the reward,
\begin{align}
  r = \mathcal{R}(\hat{\mathbf{C}},\, \mathbf{C}).
\end{align}
The group-normalized advantage of $\hat{\mathbf{C}}$ is given by
\begin{align}
  A(\hat{\mathbf{C}}) = \frac{r - \mu_\mathcal{G}}{\sigma_\mathcal{G}},
\end{align}
where $\mu_\mathcal{G}$ and $\sigma_\mathcal{G}$ are the mean and standard deviation of $r$ over $\mathcal{G}$.
The GRPO objective is
\begin{align}
  \mathcal{L}_{\text{GRPO}}(\Phi) = -\mathbb{E}\!\left[\sum_{\hat{\mathbf{C}} \in \mathcal{G}} A(\hat{\mathbf{C}}) \log \pi_\Phi(\hat{\mathbf{C}} \mid V, M)\right],
\end{align}
where we drop the KL divergence term as in \cite{wang2026time, zhang2026timelens}.

\noindent\textbf{Training Data.}
To build a training set without expensive professional annotation, we construct the dataset $\mathcal{D}_\text{GRPO}$ by distilling pseudo-labels from a frontier MLLM $\mathcal{M}_{\text{teacher}}$ over 977 videos from MLVU~\cite{zhou2024mlvu}, with durations ranging from 5 to 15 minutes.
Formally, for each source video $V$ in MLVU, $\mathcal{M}_{\text{teacher}}$ directly generates three message-edit pairs,
which yields $977 \times 3 = 2{,}931$ distillation samples in total; details of the construction procedure and dataset statistics are provided in the supplementary material.
We refer to Qwen3-VL models trained with this procedure as \textbf{MEDitor}, with MEDitor-4B and MEDitor-8B denoting the 4B and 8B variants, respectively.

%% file: sec/4_experiments.tex
\section{Experiments}
\label{sec:experiments}

\begin{table*}[tb]
  \centering
  \begin{threeparttable}
    \caption{Main results on MEDit-Bench. Temporal alignment is measured by R@$\theta$, F1@$\theta$, and mIoU; LLM Pref.\ shows the LLM-as-a-judge win rate against human reference edits, counted only when the model is preferred in both presentation orders to eliminate position-biased judgments (green: model preferred, red: human preferred). Although the LLM Pref. is relatively unreliable as an evaluation metric due to the position bias effect, we provide them here for reference. Human row shows inter-annotator performance (gray).}
    \label{tab:editnet_results}
    \begin{tabular}{lcccccccc}
      \toprule
      Model & R@0.3 & R@0.5 & R@0.7 & F1@0.3 & F1@0.5 & F1@0.7 & mIoU & LLM Pref. \\
      \midrule
      Random                & 0.030 & 0.002 & 0.000 & 0.121 & 0.054 & 0.017 & 0.112 & \winlossbar{16.7}{83.3} \\
      
      \midrule
      \multicolumn{9}{l}{\textit{Commercial models.}} \\
      GPT-5                 & \textbf{0.348} & 0.065 & 0.009 & \textbf{0.273} & \textbf{0.151} & 0.064 & \textbf{0.253} & \winlossbar{77.8}{22.2} \\
      GPT-5-mini            & 0.157 & 0.033 & 0.000 & 0.229 & 0.112 & 0.041 & 0.186 & \winlossbar{31.1}{68.9} \\
      Gemini-3-pro~\cite{geminiteam2025geminifamilyhighlycapable}   & 0.263 & 0.054 & 0.004 & 0.255 & 0.152 & \textbf{0.083} & 0.235 & \winlossbar{72.8}{27.2} \\
      Gemini-2.5-pro~\cite{geminiteam2025geminifamilyhighlycapable} & 0.144 & 0.033 & 0.006 & 0.180 & 0.102 & 0.051 & 0.152 & \winlossbar{28.3}{71.7} \\
      \midrule
      \multicolumn{9}{l}{\textit{Open-source models.}} \\
      Qwen3-VL-4B~\cite{Qwen3-VL}  & 0.032 & 0.009 & 0.002 & 0.075 & 0.036 & 0.014 & 0.119 & \winlossbar{25.2}{74.8} \\
      Gemma-4-E4B               & 0.111 & 0.017 & 0.000 & 0.156 & 0.071 & 0.024 & 0.169 & \winlossbar{30.6}{69.4} \\
      Qwen3-VL-8B~\cite{Qwen3-VL}  & 0.144 & 0.030 & 0.004 & 0.173 & 0.099 & 0.041 & 0.165 & \winlossbar{26.1}{73.9} \\
      InternVL3.5-8B~\cite{wang2025internvl3} & 0.020 & 0.000 & 0.000 & 0.106 & 0.042 & 0.011 & 0.080 & \winlossbar{7.4}{92.6} \\
      Qwen3-VL-32B~\cite{Qwen3-VL} & 0.248 & 0.070 & 0.013 & 0.234 & 0.126 & 0.061 & 0.227 & \winlossbar{40.7}{59.3} \\
      MEDitor-4B (Ours) & 0.244 & 0.069 & 0.013 & 0.250 & 0.146 & 0.071 & 0.229 & \winlossbar{52.9}{47.1} \\
      MEDitor-8B (Ours) & 0.304 & \textbf{0.100} & \textbf{0.026} & 0.258 & 0.150 & 0.061 & 0.241 & \winlossbar{54.9}{45.1} \\
       \midrule
      Human & {\color{gray} 0.432} & {\color{gray} 0.133} & {\color{gray} 0.033} & {\color{gray} 0.277} & {\color{gray} 0.184} & {\color{gray} 0.109} & {\color{gray} 0.290} & --- 
      \\
      \bottomrule
    \end{tabular}
  \end{threeparttable}
\end{table*}

\subsection{Evaluation Setup}
We evaluate a range of commercial and open-source MLLMs on MEDit-Bench in zero-shot settings, as well as MEDitor, our GRPO-based fine-tuned models (\cref{method: grpo}).
All models are assessed with temporal alignment metrics (R@$\theta$, F1@$\theta$, mIoU) and LLM-based preference (WinRate).
To further study the gap between human and model edits, we conduct a user study comparing professional human edits to those of Gemini-3-pro, yielding 1,620 evaluations in total. See the supplementary material for more details on the user study settings. We also evaluate MEDitor on classic video summarization benchmarks, \ie, SumMe~\cite{10.1007/978-3-319-10584-0_33} and TVSum~\cite{7299154}, and provide the evaluation setup and results in~\cref{sec:summarization}.

\subsection{MEDit-Bench Results}
\label{sec:results}

\noindent\textbf{Temporal alignment.}
All models outperform the Random baseline, but fall short of the human editing.
Among zero-shot models, larger commercial models generally outperform open-source ones, suggesting that general-purpose video reasoning capacity is an important factor.
Performance degrades sharply at stricter thresholds; at $\theta=0.7$, even the best model achieves only 0.026, indicating that precise cut-boundary alignment remains very challenging.
MEDitor substantially improves performance: MEDitor-8B surpasses all zero-shot open-source models and rivals commercial models, achieving the highest R@0.5 (0.100) and R@0.7 (0.026) among all evaluated models.

\noindent\textbf{LLM-based preference.}
While the win-rate metric has known reliability concerns (\cref{sec:limitations}), its ranking is broadly consistent with model capability: larger models score markedly higher than smaller ones, and MEDitor raises win rate substantially, mirroring the trend in temporal metrics.
One exception is Gemini-3-pro, whose win rate (72.8\%) appears disproportionately high relative to its temporal scores, possibly reflecting self-enhancement bias

\begin{figure*}[tb]
  \centering
  \begin{minipage}[t]{0.48\linewidth}
    \centering
    \includegraphics[width=\linewidth]{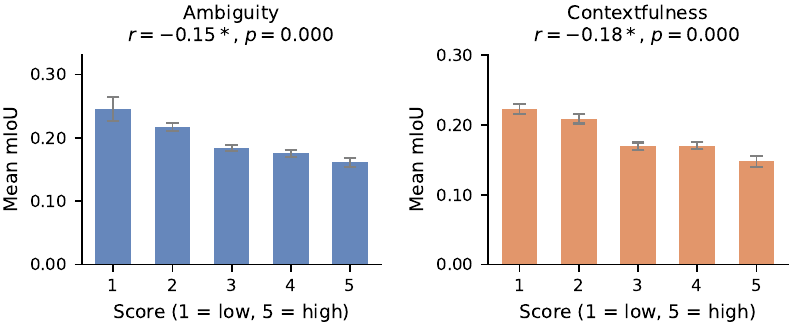}
    \caption{Mean mIoU as a function of message ambiguity (left) and contextfulness (right), aggregated across all models.
    Error bars show standard error.
    Both dimensions negatively correlate with mIoU ($^*p < 0.001$).}
    \label{fig:label_f1}
  \end{minipage}
  \hfill
  \begin{minipage}[t]{0.48\linewidth}
    \centering
    \includegraphics[width=\linewidth]{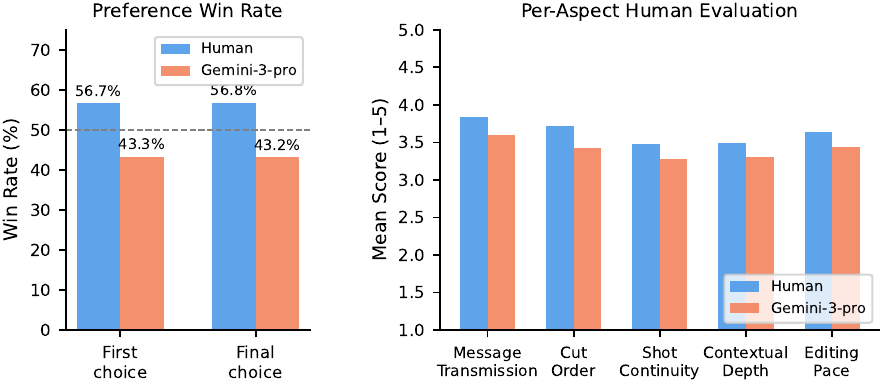}
    \caption{Human perceptual study comparing professional human edits vs.\ Gemini-3-pro outputs (1,620 evaluations).
    \emph{Left}: win rates at first and final preference choice.
    \emph{Right}: mean scores (1--5) per evaluation aspect; higher is better.}
    \label{fig:ab_test}
  \end{minipage}
\end{figure*}

\noindent\textbf{Effect of message quality.}
\Cref{fig:label_f1} shows that both ambiguity and contextfulness negatively correlate with mIoU ($r=-0.15$ and $r=-0.18$, $p<0.001$).
In particular, the most ambiguous messages (score~5) yield mIoU roughly 34\% lower than low-ambiguity ones (score~1), as highly open-ended narratives admit many valid interpretations.
Contextful messages pose a distinct challenge as well, requiring models to infer prerequisite background content rather than select scenes that explicitly match the message.
These results establish message quality as a meaningful stratification factor for evaluation.

\noindent\textbf{User Study results. }
Three independent evaluators assessed each edit on five aspects: \textit{(A) message transmission} (how well the edit conveys the intended narrative), \textit{(B) cut order} (whether the ordering of shots effectively serves the intended message and narrative), \textit{(C) shot continuity} (whether transitions between cuts are smooth), \textit{(D) contextual depth} (whether sufficient context is provided to make the narrative meaningful), and \textit{(E) narrative pacing} (whether the timing and rhythm of cuts feel appropriate).
They also provided A/B preference judgments before and after the questionnaire.
\Cref{fig:ab_test} summarizes the results.
Human edits were preferred in 56.7\% of first-choice evaluations and 56.8\% of final-choice evaluations, a preference that is consistent across both judgment stages. This means that the awareness of specific evaluation aspects does not affect the overall preference.
Across all five evaluated aspects, human edits consistently score higher than the model's edits.
The largest gaps appear in (B) \emph{cut order} and (A) \emph{message transmission}, indicating that models particularly struggle to construct an ordering of shots that conveys the narrative.
The gap is smallest for (C) \emph{shot continuity}, suggesting that models achieve acceptable temporal transitions even when the overall narrative is weaker.

\begin{figure*}[t]
  \centering
  \includegraphics[width=\textwidth]{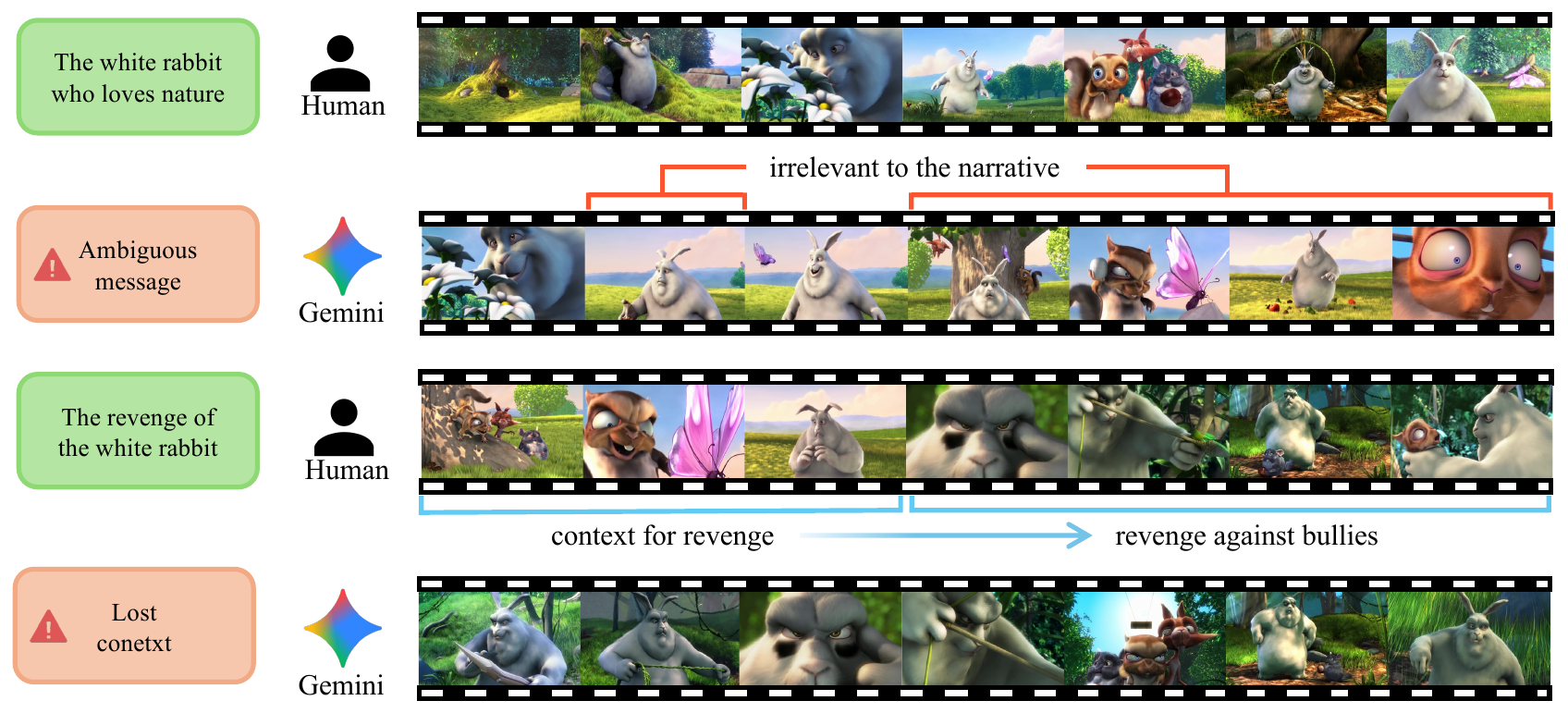}
  \caption{Qualitative comparison of human and Gemini-3-pro edits on two cases where all evaluators unanimously preferred the human edit.
  Each pair shows the human edit (top strip) and the Gemini edit (bottom strip) alongside the diagnosed failure mode.
  \emph{Top} (``The white rabbit who loves nature''): the abstract message causes Gemini to select clips irrelevant to the intended narrative.
  \emph{Bottom} (``The revenge of the white rabbit''): Gemini omits the contextual buildup needed to make the reversal meaningful.}
  \label{fig:qualitative}
\end{figure*}

\input{sec/4.2_qualitative}

\input{sec/4.1_ablation}

%% file: sec/4.2_qualitative.tex
\subsection{Qualitative Analysis}
\label{sec:qual_analy}

To illustrate the nature of the gap between humans and models, we present qualitative comparisons on two video-message pairs for which all three human evaluators unanimously preferred the human edit and the score gap was especially pronounced (\cref{fig:qualitative}).

\noindent\textbf{Sample 1 (message: \textit{``The white rabbit who loves nature''}).}
The message lacks references to concrete objects, making the cut selection criterion inherently ambiguous.
The model (Gemini-3-pro) selects clips mostly irrelevant to the nuance of the message, failing to ground the abstract message in visual content.
Evaluators rated the model particularly low on message transmission and cut order (Human/model: 4.33/2.00 and 4.00/2.00, respectively).
We observe that models tend to handle messages containing explicit object names more reliably, suggesting that message concreteness is a key factor for good edits.

\noindent\textbf{Sample 2 (message: \textit{``The revenge of the white rabbit''}).}
This message implies a causal arc from victimization to retaliation.
The human editor includes contextual buildup, \ie, scenes establishing the bullying dynamic, before the counterattack, making the eventual reversal comprehensible and emotionally resonant.
The model cuts directly to the revenge without sufficient setup, losing the narrative context that gives the revenge its meaning.
The largest gap appears in cut order (Human/model: 4.33/1.67), whereas shot continuity is rated \emph{higher} for the model edit (4.00 vs.\ 3.33), demonstrating that temporal smoothness alone does not imply narrative coherence.

%% file: sec/4.1_ablation.tex
\subsection{Generalization to Video Summarization}
\label{sec:summarization}



\begin{table}[tb]
  \centering
  \small
  \begin{threeparttable}
    \caption{Zero-shot performance on video summarization benchmarks. \textbf{Bold} and \underline{underline} denote the best and second-best scores per metric. \emph{Human} row: per video, each annotator is scored against the consensus of the remaining annotators (SumMe) or against every other annotator individually (TVSum), then averaged over annotators.}
    \label{tab:ablation_summary}
    \begin{tabular}{lcccc}
      \toprule
      \multirow{2}{*}{Method} & \multicolumn{2}{c}{SumMe} & \multicolumn{2}{c}{TVSum} \\
      \cmidrule(lr){2-3} \cmidrule(lr){4-5}
      & $\tau$ & $\rho$ & $\tau$ & $\rho$ \\
      \midrule
      Random & 0.000 & 0.000 & 0.000 & 0.000 \\
      Human  & 0.296 & 0.329 & 0.177 & 0.204 \\
      \midrule
      \multicolumn{5}{l}{\textit{Vision-only methods.}} \\
      CSNet~\cite{CSNet}      & --- & --- & 0.025 & 0.034 \\
      CASUM~\cite{CASUM}      & 0.063 & 0.084 & 0.160 & 0.210 \\
      DSAVS~\cite{DSAVS}      & --- & --- & 0.080 & 0.087 \\
      DRDSN~\cite{DRDSN}      & 0.047 & 0.048 & 0.020 & 0.026 \\
      RSGN$_u$ \cite{RSGN_u} & 0.071 & 0.073 & 0.048 & 0.052 \\
      RSSUM~\cite{RSSUM}      & 0.007 & 0.015 & 0.080 & 0.106 \\
      CTVSum~\cite{pang2023contrastive} & 0.036 & 0.044 & 0.161 & 0.212 \\
      \midrule
      \multicolumn{5}{l}{\textit{Vision-language methods.}} \\
      LGSSVS~\cite{LGSSVS}     & 0.102 & 0.148 & 0.133 & 0.174 \\
      LfVS (zeroshot)~\cite{argaw2024scaling} & 0.125 & 0.148 & 0.151 & 0.182 \\
      Ge et al.~\cite{10.1007/978-981-96-5815-2_6}\ & 0.157 & 0.182 & \underline{0.184} & 0.213 \\
      SSPVS~\cite{li2023progressive} & 0.154 & \underline{0.207} & 0.151 & 0.199 \\
      Qwen3-VL-4B~\cite{Qwen3-VL}        & 0.138 & 0.161 & 0.081 & 0.092 \\
      Qwen3-VL-8B~\cite{Qwen3-VL}        & \underline{0.160} & 0.186 & 0.089 & 0.101 \\
      MEDitor-4B (Ours) & 0.149 & 0.172 & \textbf{0.193} & \textbf{0.219} \\
      MEDitor-8B (Ours) & \textbf{0.218} & \textbf{0.254} & \textbf{0.193} & \underline{0.218} \\
      \bottomrule
    \end{tabular}
  \end{threeparttable}
\end{table}
\noindent\textbf{Setup.}
We probe whether the capabilities acquired through GRPO training transfer to , without any additional fine-tuning.
The prompt format and model weights are kept identical to the message-driven editing task, replacing only the message $M$.
Because the model's output is sensitive to the phrasing of $M$ (different messages select different cuts), we do not rely on a single message but \emph{ensemble} a small set of generic highlight messages (``critical'', ``exciting'', and ``memorable moments''), running the model once per message and averaging their per-frame selection masks into a graded, phrasing-robust per-frame importance.
We then compute Kendall's $\tau$ and Spearman's $\rho$ between this importance and the human ground truth by following~\cite{8954229}. Following \cite{son2024csta}, for SumMe, the reference is the consensus selection frequency (the fraction of annotators selecting each frame); for TVSum, the coefficients are computed for each annotator separately and then averaged.

\noindent\textbf{Results.}
Although neither benchmark is seen during training, the editing capability transfers zero-shot. MEDitor-8B attains the best SumMe scores ($\tau$/$\rho$ 0.218/0.254) and both MEDitor models lead on TVSum (0.193/0.219), surpassing prior unsupervised and vision-language methods on both benchmarks. The improvement over the untrained Qwen3-VL backbones is largest on TVSum, where $\tau$ climbs from 0.081 (4B) and 0.089 (8B) to 0.193, confirming that the transfer is driven by GRPO training rather than the base model. On TVSum our models even exceed the leave-one-out human agreement (0.177/0.204): they match the annotator consensus more closely than individual annotators match one another, reflecting the well-known annotator disagreement on this benchmark. On SumMe they remain below the higher human agreement (0.296/0.329), and model scale helps there (8B $>$ 4B) while making little difference on TVSum.

%% file: sec/7_limitations.tex
\section{Limitations}
\label{sec:limitations}

MEDit-Bench focuses on cut-only editing, whereas professional video production also involves titles and captions, color grading, and motion effects. 
Cut selection is nonetheless the most fundamental step that drives all subsequent editing decisions, making it a natural starting point for benchmarking message-driven video editing.
On the evaluation side, the LLM-as-a-judge protocol has known limitations: the judge may lack sensitivity to narrative coherence and professional editing craft, and is susceptible to position bias and self-enhancement bias despite our order-swapping mitigation; we provide a detailed analysis of these limitations in the supplementary material.
Furthermore, relying on commercial models such as Gemini or GPT as evaluation judges raises a reproducibility concern: these models are subject to deprecation and silent version updates, meaning that the same evaluation prompt may yield different results over time and cannot serve as a fixed, reproducible metric.

%% file: sec/5_conclusion.tex
\section{Conclusion}
\label{sec:conclusion}

We introduced MEDit-Bench, a dataset and benchmark for message-driven video editing, pairing long-form videos with multiple natural-language editing messages and professionally produced edits.
Our benchmark scores edits automatically with temporal alignment metrics (R@$\theta$, F1@$\theta$, mIoU), quantifying the structural quality of cut selection and placement.
We also explored automatic semantic evaluation via LLM-as-a-judge, but found it unreliable for this task due to severe position bias, a cautionary result for automatic narrative-quality evaluation.
Experiments show that precise cut placement and narrative coherence remain open challenges, with message ambiguity and contextfulness stratifying model performance further.
A complementary human perceptual study confirms this gap, with professional edits consistently preferred over model outputs across all evaluated narrative aspects.
We hope MEDit-Bench provides a rigorous foundation for advancing message-aware video editing research.

%% file: sec/8_supplementary.tex
This supplementary material provides (A)~dataset collection details including statistics, editor recruitment, and message creation,
(B)~implementation details including GRPO training data construction,
(C)~user study protocol details,
(D)~LLM-as-judge limitations,
(E)~full inference prompts, and (F)~representative dataset examples.

\appendix

\section{Dataset Collection Details}
\label{sec:collection}

\subsection{Video Selection}
\label{sec:video_selection}

We selected 60 videos from LongVideoBench~\cite{wu2024longvideobenchbenchmarklongcontextinterleaved} and VideoMME~\cite{fu2025video}, applying the following criteria.

\noindent\textbf{Domain diversity.}
We aimed for a balanced distribution across eight categories to ensure the benchmark covers a wide range of visual styles and content types.
\Cref{tab:domain_dist} shows the resulting distribution.

\begin{table}[h]
  \centering
  \small
  \caption{Domain distribution of MEDit-Bench source videos.}
  \label{tab:domain_dist}
  \begin{tabular}{lc}
    \toprule
    \textbf{Category} & \textbf{\# Videos} \\
    \midrule
    Vlog        & 14 \\
    Travel      & 13 \\
    Movie       & 9  \\
    Animation   & 6  \\
    Sports      & 5  \\
    Documentary & 5  \\
    Food        & 4  \\
    Cooking     & 4  \\
    \midrule
    \textbf{Total} & \textbf{60} \\
    \bottomrule
  \end{tabular}
\end{table}

\noindent\textbf{Message potential.}
We selected videos with visually rich and varied content, enabling the creation of multiple distinct editing messages per video.

\noindent\textbf{Visual editability.}
We excluded videos in which the primary information is conveyed through speech rather than visual content (\eg, news broadcasts where spoken text is essential for understanding), ensuring that all editing decisions can be made based on visual content alone.

\subsection{Dataset Statistics}
\label{sec:dataset_stats}

\begin{figure*}[h]
  \centering
  \includegraphics[width=\linewidth]{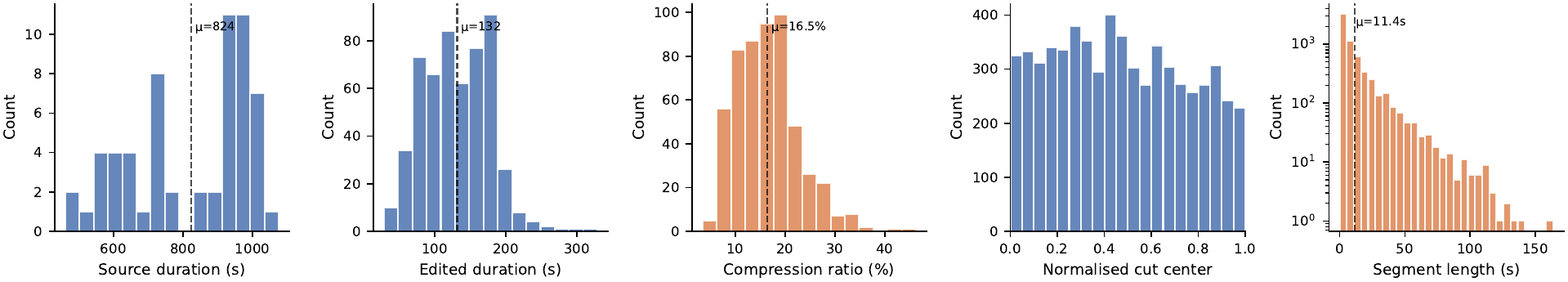}
  \caption{Statistics of the MEDit-Bench dataset.}
  \label{fig:dataset_statistics}
\end{figure*}

\Cref{fig:dataset_statistics} summarizes the statistics of MEDit-Bench.
Source videos range from 461 to 1,076 seconds (824 seconds on average), while edited videos range from 28 to 329 seconds (132 seconds on average), with an average compression ratio of 16.5\%.
Cuts are distributed almost uniformly over time, indicating minimal temporal bias.
MEDit-Bench contains 6,231 cuts in total; the average cut length is 11.40 seconds (median 5.09 seconds).

\noindent\textbf{Shot reordering.}
To characterize the extent to which editors deviate from chronological ordering, we count cuts whose start timestamp precedes that of the immediately preceding cut in the edit sequence.
\Cref{tab:reorder} shows the reordering rates for all evaluated models and the human reference.
Among professional edits, 26.1\% of edits and 5.3\% of cuts involve at least one temporal reversal, whereas current AI models produce near-zero reordering rates (0--2\%).
This gap highlights that intentional shot reordering, an important tool in professional narrative editing, is largely absent from model outputs, representing a qualitative gap beyond what temporal alignment metrics alone can capture.

\begin{table}[h]
  \centering
  \caption{Shot reordering rates for AI models and human reference editors.
           ``Edits'' counts edits with at least one temporal reversal;
           ``Cuts'' counts individual cuts placed before the preceding cut.}
  \label{tab:reorder}
  \resizebox{\columnwidth}{!}{%
  \begin{tabular}{lcc}
    \toprule
    \textbf{Model} & \textbf{Edits w/ reorder} & \textbf{Cuts (reorder)} \\
    \midrule
    GPT-5              & 0/180 (0.0\%)  & 0/1542 (0.0\%)  \\
    GPT-5-mini         & 3/180 (1.7\%)  & 3/1106 (0.3\%)  \\
    Gemini-3-pro       & 0/180 (0.0\%)  & 0/2025 (0.0\%)  \\
    Gemini-2.5-pro     & 0/180 (0.0\%)  & 0/1582 (0.0\%)  \\
    \midrule
    Qwen3-VL-4B        & 14/176 (8.0\%) & 124/15939 (0.8\%) \\
    Gemma-4-E4B        & 1/180 (0.6\%)  & 1/934 (0.1\%)   \\
    Qwen3-VL-8B        & 0/180 (0.0\%)  & 0/2304 (0.0\%)  \\
    InternVL3.5-8B     & 0/180 (0.0\%)  & 0/1472 (0.0\%)  \\
    Qwen3-VL-32B       & 0/180 (0.0\%)  & 0/2730 (0.0\%)  \\
    MEDitor-4B         & 2/180 (1.1\%)  & 2/2030 (0.1\%)  \\
    MEDitor-8B         & 4/180 (2.2\%)  & 5/1503 (0.3\%)  \\
    \midrule
    Human (reference)  & \textbf{141/540 (26.1\%)} & \textbf{329/6231 (5.3\%)} \\
    \bottomrule
  \end{tabular}}
\end{table}

\subsection{Editor Recruitment and Compensation}
\label{sec:editors}

Editors were recruited through a crowdsourcing platform via an open call.
Applicants were not filtered by formal credentials at the application stage; instead, editors were selected from among applicants based on their platform rating and demonstrated editing experience.
Each editor was compensated at \textbf{10,000-15,000 JPY per source video}, which covers producing three edited videos (one per message) along with the required deliverables.

\subsection{Editor Instructions}
\label{sec:editor_instructions}

Each editor received a source video (10--15 minutes) and three editing messages, and was asked to produce one edited video per message.
Editors are instructed to (i)~trim the source to a 1--3 minute summary using only cut operations, (ii)~reorder shots freely, (iii)~ignore the audio and base their decisions solely on the visual content, and (iv)~provide a rationale for every cut to discourage careless cuts.
Although the shot ordering and rationales are not used in our benchmark, we release them to the community for future research.

For each edited video, editors were required to submit three deliverables:
(i)~an exported video file (mp4),
(ii)~the editing project file (Adobe Premiere Pro or equivalent),
and (iii)~a rationale memo (one line per cut) explaining why each scene was selected.

\subsection{Data Availability}
\label{sec:data_availability}

The released dataset consists of (i) YouTube video IDs, (ii) the editing messages in text form, and (iii) the cut timestamps for each professional edit, extracted from the editors' project files (Adobe Premiere Pro or equivalent) to ensure frame-level precision.
The source videos originate from LongVideoBench~\cite{wu2024longvideobenchbenchmarklongcontextinterleaved} and VideoMME~\cite{fu2025video}, both of which are publicly available datasets, so the underlying footage can be freely accessed.

\subsection{Quality Control}
\label{sec:quality_control}

All submitted edits were manually reviewed by the authors.
For edits with an unusually low number of cuts or those that appeared superficial, the editing rationale memo was cross-checked against the actual cut sequence to verify that each cut was intentional and consistent with the stated intent.
Submissions deemed to fall below an acceptable quality level were returned to the editor with a request for resubmission.
No strictly formalized acceptance criteria were applied; quality judgments were made by the authors based on overall consistency between the edit, the message, and the rationale.

\subsection{Message Creation Process}
\label{sec:message_creation}

After watching each source video in full, the authors created three editing messages per video.
Messages were designed to satisfy the following properties.

\noindent\textbf{Visual communicability.}
Each message describes a narrative that can be conveyed through visual content alone, without relying on audio or spoken dialogue, consistent with the visual-only editing constraint imposed on editors.

\noindent\textbf{Diversity of intent.}
The three messages per video were designed to evoke distinctly different impressions or narratives from the same footage, ensuring that different messages yield substantially different edits.

\noindent\textbf{Varied difficulty.}
The message set deliberately spans a range of difficulty levels.
This includes straightforward messages with clear visual anchors, ambiguous messages open to multiple equally valid interpretations, and context-dependent messages that require introductory shots to make the narrative meaningful to viewers.
This design is reflected in the ambiguity and contextfulness annotations described in the main paper (Sec. 3.3).

\section{Implementation Details}
\label{sec:impl}

\subsection{Open-Source Models}
\label{sec:impl_open_source}

\noindent\textbf{Qwen3-VL.}
We evaluate Qwen3-VL in 4B, 8B, and 32B parameter variants.
Videos are sampled at 2 FPS with a maximum of 2,048 frames.
Spatial resolution is controlled by a per-frame lower bound of \texttt{min\_pixels\,=\,65,536} and a cross-frame upper bound of \texttt{total\_pixels\,=\,20,971,520}, which allows the model to adaptively trade frame count against per-frame resolution based on video length.
The maximum number of generated tokens is set to 1024.

\noindent\textbf{Gemma-4.}
We evaluate Gemma-4-E4B using its native video processing pipeline.
Thinking mode is disabled (\texttt{enable\_thinking\,=\,False}) and the maximum number of generated tokens is set to 1,024.

\noindent\textbf{InternVL3.5.}
We evaluate InternVL3.5-8B by uniformly sampling 32 frames from each video.
Each frame is resized to $448{\times}448$ and treated as a single tile (\texttt{max\_num\,=\,1}, no dynamic sub-image splitting).
The maximum number of generated tokens is set to 1024.

\subsection{Commercial Models}
\label{sec:impl_commercial}

\noindent\textbf{GPT-5.}
Following~\cite{zhang2026timelens}, we sample frames at 1 FPS and prepend a textual timestamp label (\eg, \texttt{Frame at 2.5s:}) to each frame.
For videos shorter than 50 seconds all frames are passed directly; for videos of 50--80 seconds we uniformly downsample to 50 frames; for videos longer than 80 seconds we sample at 1 FPS and arrange every 4 consecutive frames into a $2{\times}2$ grid image.
For GPT-5, which is a reasoning model, we use the default value for the \texttt{reasoning\_effort} parameter.
GPT-5-mini uses the same pipeline.

\noindent\textbf{Gemini.}
Source videos are uploaded to the Google File API and passed directly to the model via the \texttt{file\_data} interface, delegating all frame sampling and resolution control to the model's native video processing.
We evaluate both Gemini-2.5-Pro and Gemini-3-Pro using this pipeline.

\subsection{GRPO Training Data Construction}
\label{sec:impl_grpo_data}

We use GPT-5 as the teacher model $\mathcal{M}_\text{teacher}$ to generate pseudo-labeled (message, cut sequence) pairs over 977 source videos drawn from the MLVU benchmark~\cite{zhou2024mlvu}, with durations ranging from 5 to 15 minutes.
For each video, the teacher generates three message-edit pairs, yielding $977 \times 3 = 2{,}931$ training samples in total.

\begin{figure*}[h]
  \centering
  \includegraphics[width=\textwidth]{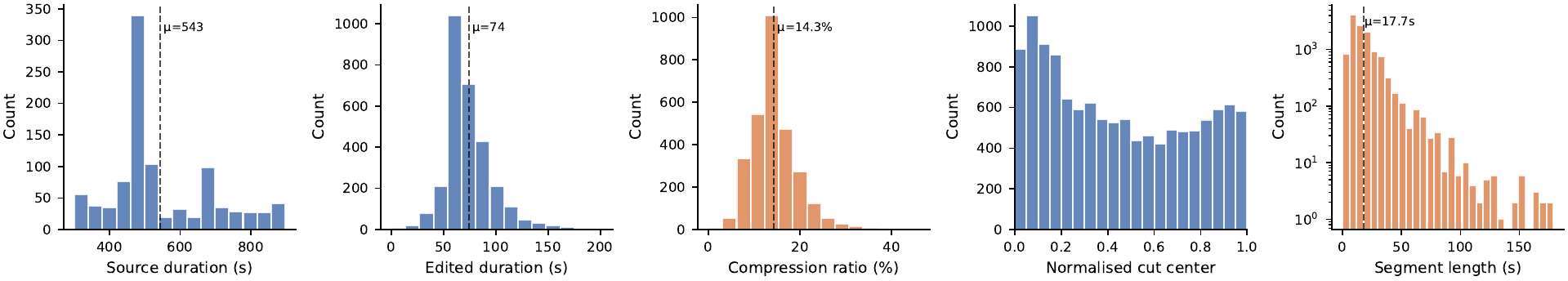}
  \caption{%
    Distribution statistics of the MEDitor (GRPO) training data.
    From left to right: source video duration (s), generated edit duration (s), compression ratio (\%), normalised cut center, and segment length (s, log scale).
    Dashed lines mark the mean.
  }
  \label{fig:grpo_stats}
\end{figure*}

\noindent\textbf{Dataset statistics.}
\Cref{tab:grpo_stats} and \Cref{fig:grpo_stats} summarize the key statistics of the resulting training set.
Source footage averages 543 seconds (9.1 min), and the generated edits average 74 seconds with a compression ratio of approximately 14.3\%.
Each generated edit selects on average 4.2 cuts, with an average segment length of 17.7 seconds.

\begin{table}[h]
  \centering
  \caption{Statistics of MEDit-Bench and MEDitor (GRPO) training data.}
  \label{tab:grpo_stats}
  \resizebox{\columnwidth}{!}{%
  \begin{tabular}{lccc}
    \toprule
    & & \textbf{MEDit-Bench} & \textbf{MEDitor (GRPO)} \\
    \midrule
    \multirow{2}{*}{Source}
      & Mean   & 824 s  & 543 s  \\
      & Median & 905 s  & 490 s  \\
    \midrule
    \multirow{2}{*}{Edit duration}
      & Mean   & 132 s  & 74 s   \\
      & Median & 130 s  & 69 s   \\
    \midrule
    \multirow{2}{*}{Cuts per edit}
      & Mean   & 11.5   & 4.2    \\
      & Median & 6.0    & 4.0    \\
    \midrule
    \multirow{2}{*}{Compression ratio}
      & Mean   & 16.5\% & 14.3\% \\
      & Median & 16.2\% & 13.7\% \\
    \midrule
    \multirow{2}{*}{Segment length}
      & Mean   & 11.4 s & 17.7 s \\
      & Median & 5.1 s  & 14.0 s \\
    \bottomrule
  \end{tabular}}
\end{table}

\Cref{fig:grpo_stats} shows the distributions of source duration, edited duration, compression ratio, normalised cut center, and segment length across the 977 training videos and 2,931 generated edits.

\noindent\textbf{Generation prompt.}
The prompt used to generate pseudo-labels is shown in \Cref{sec:prompt_pseudo}.

\subsection{MEDitor Training and Inference}
\label{sec:impl_meditor}

\noindent\textbf{Training.}
We initialize MEDitor directly from Qwen3-VL-4B/8B-Instruct and train it with GRPO on the 2,931 distilled (message, cut sequence) pairs.
For every sample we draw $G=4$ rollouts at temperature~1.0, normalize the advantages within the group, and drop the KL term ($\beta=0$).
The vision encoder is frozen while the language model and the vision-language merger are trained, with a constant learning rate of $1{\times}10^{-6}$, global batch size~64 for half an epoch, beyond which the training reward starts to plateau.
Videos are decoded at 2\,fps and capped at 448 frames with a total visual-token budget of 14{,}336 (per-frame floor 64).

\noindent\textbf{Inference.}
At test time MEDitor uses greedy decoding (no sampling) with the same video sampling budget as training.
Given a video and message, it emits the plain seconds list specified by the prompt, which we parse into the predicted cut set; no post-processing or re-ranking is applied.

\subsection{Human Performance Evaluation}
\label{sec:human_eval}

The \textit{Human} row in the Table 3 reports inter-annotator performance under a leave-one-out protocol.
For each (video, message) pair, one of the three editors is designated as the predictor and the remaining two as references.
All temporal alignment metrics (R@$\theta$, F1@$\theta$, mIoU) are computed between the predictor's edit and each reference edit separately, then averaged over the two references.
This procedure is repeated for all three choices of predictor editor, and the final score is the average over all three combinations and all (video, message) pairs in MEDit-Bench.

\section{Ablation Studies}
\label{sec:ablation}

\subsection{Reward Function Ablation}
\label{sec:ablation_reward}

We ablate the GRPO reward on the MEDit-Bench to isolate the contribution of each term in our combined reward $\mathcal{R} = \mathcal{R}_{\mathrm{cov}} + \mathcal{R}_{\mathrm{F1}}$.
All three variants train the base Qwen3-VL-4B from scratch under identical settings (GPT-5 pseudo-labels, half an epoch, global batch~64) and differ only in the reward: coverage only ($\mathcal{R}_{\mathrm{cov}}$), cut-matching only ($\mathcal{R}_{\mathrm{F1}}$), and their equal-weight combination.

\begin{table}[h]
  \centering
  \caption{Reward ablation on the MEDit-Bench. mIoU is union-mask coverage; F1@$\tau$ is optimal cut-match F1 at IoU~$\tau$. \textbf{Bold} = best among the three rewards.}
  \label{tab:reward_ablation}
  \resizebox{\columnwidth}{!}{%
  \begin{tabular}{lccccc}
    \toprule
    Reward & mIoU & F1@.3 & F1@.5 & F1@.7 & Cuts/edit \\
    \midrule
    Raw (no training)                  & 11.90 & 7.48  & 3.60  & 1.35  & 2.0 \\
    \midrule
    $\mathcal{R}_{\mathrm{cov}}$ only   & 24.01 & 12.18 & 5.68  & 3.72  & 1.8 \\
    $\mathcal{R}_{\mathrm{F1}}$ only    & 22.80 & 25.26 & 13.64 & 5.91  & 8.5 \\
    Combined (ours)                    & \textbf{24.73} & \textbf{25.36} & \textbf{15.26} & \textbf{7.48} & 8.2 \\
    \midrule
    Human (ceiling)                    & 28.97 & 27.69 & 18.44 & 10.88 & 11.5 \\
    \bottomrule
  \end{tabular}}
\end{table}

\noindent\textbf{Findings.}
The combined reward is best on all four alignment metrics, and the two terms are complementary rather than redundant.
Coverage alone ($\mathcal{R}_{\mathrm{cov}}$) is blind to how the timeline is partitioned, so the policy collapses to a single fat span (1.8 cuts vs.\ the human 11.5), matches at most a couple of the human moments, and craters F1 (F1@.5 $=5.68$).
Cut-matching alone ($\mathcal{R}_{\mathrm{F1}}$) restores the cut count (8.5 cuts) and more than doubles F1, but is slightly weaker on coverage.
Combining them retains both properties: the right number of well-placed cuts and the highest coverage.

\subsection{Message Ensembling for Video Summarization}
\label{sec:ablation_msgens}

Our video-summarization evaluation runs the model with three generic messages (``critical'', ``exciting'', and ``memorable'' moments) and averages their per-frame selection masks into a single soft importance, which is then scored once.
We emphasize that AVG3 averages the model's \emph{predictions}, not the per-message metrics: it forms one averaged prediction and computes a single $\tau$/$\rho$ from it, rather than scoring each message separately and averaging the three coefficients.
\Cref{tab:msg_ensemble} decomposes this ensemble against each constituent message, where every single-message column is scored on that message's own prediction.

\begin{table}[h]
  \centering
  \caption{Message-ensembling ablation for video summarization (Kendall $\tau$). Single-message columns score each message's prediction on its own; AVG3 averages the three predictions (per-frame masks) into one importance and scores that single averaged prediction.}
  \label{tab:msg_ensemble}
  \resizebox{\columnwidth}{!}{%
  \begin{tabular}{llcccc}
    \toprule
    Model & Data & critical & exciting & memorable & AVG3 \\
    \midrule
    \multirow{2}{*}{Qwen3-VL-4B}
      & SumMe & 0.121 & 0.136 & 0.107 & \textbf{0.138} \\
      & TVSum & 0.056 & \textbf{0.092} & 0.064 & 0.081 \\
    \midrule
    \multirow{2}{*}{Qwen3-VL-8B}
      & SumMe & 0.106 & \textbf{0.163} & 0.146 & 0.160 \\
      & TVSum & \textbf{0.092} & 0.076 & 0.052 & 0.089 \\
    \midrule
    \multirow{2}{*}{MEDitor-4B}
      & SumMe & \textbf{0.159} & 0.132 & 0.108 & 0.149 \\
      & TVSum & 0.181 & 0.189 & 0.168 & \textbf{0.193} \\
    \midrule
    \multirow{2}{*}{MEDitor-8B}
      & SumMe & 0.198 & 0.197 & 0.217 & \textbf{0.218} \\
      & TVSum & 0.176 & 0.162 & 0.150 & \textbf{0.193} \\
    \bottomrule
  \end{tabular}}
\end{table}

\noindent\textbf{Findings.}
No single message is consistently best: the top-scoring message changes across models and datasets (e.g., ``critical'' for MEDitor-4B on SumMe but ``exciting'' on TVSum).
Averaging the three avoids this brittle choice while matching or exceeding the best single message; for both MEDitor models on TVSum, AVG3 strictly beats every individual message (e.g., MEDitor-8B: 0.193 vs.\ at most 0.176).

\section{User Study Details}
\label{sec:user_study}

\subsection{Study Design}
\label{sec:study_design}

We conducted a human perceptual study comparing professional human edits
to Gemini-3-pro outputs on \textbf{90 (video, message) pairs} sampled
from MEDit-Bench.
For each pair, \textbf{three independent evaluators} watched both edits
side-by-side (Video~A: human edit; Video~B: Gemini-3-pro
output) and assessed them along five aspects.
Evaluators first provided an \emph{immediate} preference choice, then
scored each aspect on a 5-point Likert scale, and finally gave a
\emph{considered} preference choice after reviewing all five scores.
This yields $90{\times}3{\times}2 = 540$ preference judgments and
$90{\times}3{\times}5 = 1{,}350$ aspect ratings
(\textbf{1{,}620 evaluations} in total).

\noindent\textbf{Participants.}
Evaluators were recruited through a crowdsourcing platform with no restriction on gender or age; participants were drawn randomly from the available pool.
Each evaluator was compensated at approximately \textbf{300 JPY per (video, message) pair evaluated}.

\subsection{Evaluation Aspects}
\label{sec:eval_aspects}

Evaluators rated each edit on five aspects adapted from film editing
practice (higher = better):

\begin{enumerate}[label=\textbf{(\Alph*)}]
  \item \textbf{Message Transmission.}
    Does the edit successfully communicate the intended message?
  \item \textbf{Cut Order.}
    Is the ordering of shots logical for conveying the narrative?
  \item \textbf{Shot Continuity.}
    Do transitions between cuts feel natural and visually coherent?
  \item \textbf{Contextual Depth.}
    Does the edit provide sufficient context for the narrative?
  \item \textbf{Editing Pace.}
    Is the pacing appropriate for the intended message?
\end{enumerate}

\section{LLM-as-Judge Limitations}
\label{sec:llm_limitations}

\begin{figure*}[t]
  \centering
  \raisebox{0.25cm}{\includegraphics[height=3.8cm]{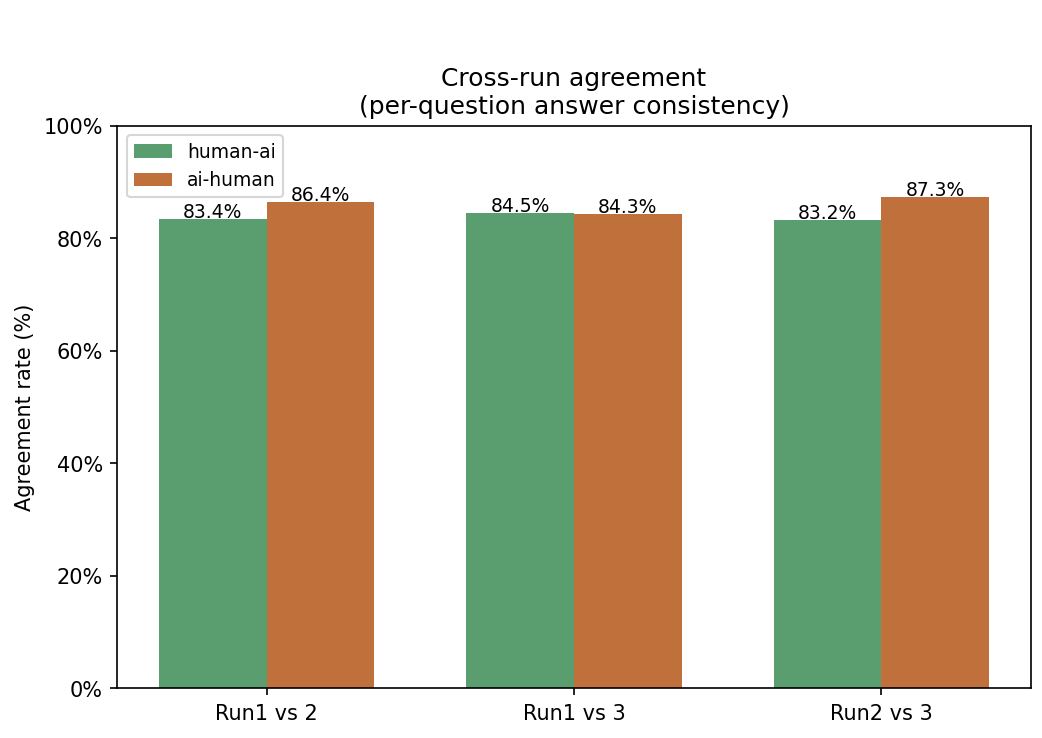}}
  \hfill
  \includegraphics[height=3.8cm]{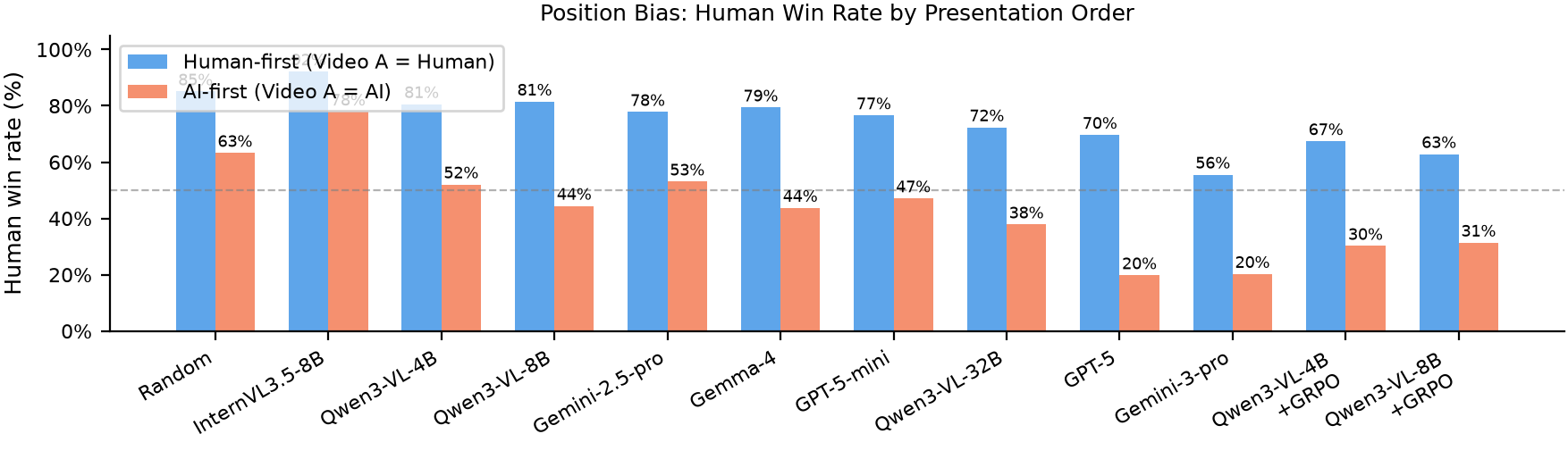}
  \caption{%
    LLM-as-judge limitations.
    \textbf{Left}: cross-run pairwise agreement for Gemini-3-pro across
    three independent runs in both conditions; 83--87\% agreement
    confirms the judge is stable across repeated trials.
    \textbf{Right}: human vs.\ AI win rate per model under human-ai
    (Video~A\,=\,human) and ai-human (Video~A\,=\,AI) conditions;
    win rates shift dramatically with presentation order, a consistent
    position bias across all models.
  }
  \label{fig:position_bias}
\end{figure*}

\noindent\textbf{Position bias.}
LLM judges tend to favor whichever video appears in position~A,
independent of content quality.
We quantify this on our Gemini-3-pro A/B evaluation (Gemini-3.5-flash
as judge): when the human edit is Video~A (\emph{human-ai} condition),
the judge selects the human edit \textbf{55.6\%} of the time; when
positions are swapped (\emph{ai-human} condition), the human win rate
drops to \textbf{20.2\%}, a swing of \textbf{35.4\,pp} driven purely
by presentation order.
Decomposing this swing, the video-A preference contributes
$+17.7\,\text{pp}$ and the residual AI content advantage contributes
$+12.1\,\text{pp}$.
As shown in \Cref{fig:position_bias} (right), this position effect is
consistent across all evaluated models.
We mitigate it by averaging judgments from both presentation orders;
the resulting \emph{position-averaged} win rates (Human~37.9\%,
AI~62.1\%) are the numbers reported in the main paper.

\noindent\textbf{Run-to-run stability.}
Despite the large position effect, repeated runs under the \emph{same}
presentation order are highly consistent.
Pairwise agreement across three independent runs ranges from
\textbf{83\% to 87\%} (\Cref{fig:position_bias}, left), confirming
that the judge is stable and that additional trials under identical
conditions do not alter the conclusions.

\noindent\textbf{Correlation with human judgment.}
To examine how well LLM preferences align with human perception, we compare LLM-as-a-judge outcomes with human judgments on the same 540 triples in $\mathcal{D}$ (Gemini-3-pro \textit{vs}.\ human reference edits).
For human preference, we use the user study results.
For LLM preference, we fix the presentation order (human-first, \ie, $\mathcal{J}(\mathbf{C}, \hat{\mathbf{C}})$) and query the judge three times independently, then aggregate via majority voting.\footnote{Due to the stochastic nature of LLM sampling, repeated runs under identical conditions can yield different individual judgments.}

The pair-wise agreement rate is \textbf{59.9\%} (chance baseline = 50\%), with Cohen's $\kappa = 0.18$, indicating only fair agreement above chance.
Agreement rises to \textbf{66.1\%} when restricted to triples where all three runs are unanimous for both the human and LLM judge, suggesting the judge is most consistent when the quality difference is clear-cut.
Treating judgments as continuous probabilities (fraction of runs preferring the model), Pearson's $r = 0.185$ and Spearman's $\rho = 0.167$ (both $p < 0.001$) confirm a statistically significant but modest monotonic association.
Overall, these results show that the LLM judge captures a weak signal of human preference, but the low $\kappa$ value underscores that it should not be treated as a reliable substitute for human evaluation.

\section{Prompts}
\label{sec:prompts}

\subsection{Zero-shot Editing Prompt}
\label{sec:prompt_zeroshot}

All MLLM baselines (Qwen3-VL-8B, Gemma-4, InternVL3.5-8B, GPT-5,
Gemini-2.5-pro, Gemini-3-pro) receive the identical prompt below,
where \texttt{\{message\}} is replaced by the actual editing message.

\begin{promptbox}{Zero-shot Editing Prompt}
You are a professional video editor. Review the
provided video and select the key temporal 
highlights that best convey the following message to viewers:

Message: "{message}"

Respond strictly with JSON using this schema:
{
  "segments": [
    ["mm:ss.ff", "mm:ss.ff"],
    ...
  ]
}

Rules:
- Select segments that best communicate the message: "{message}"
- The total duration of all selected segments combined should be approximately 1 to 3 minutes
- Choose the most visually compelling and relevant moments
- Ensure all timestamps are within the video length
- Match the precision pattern "mm:ss.ff" (minutes, seconds, hundredths)
- Do not include descriptions, titles, or any extra commentary
- Return only valid JSON with no surrounding text

Example:
{"segments": [["00:15.46", "00:19.42"], ["01:16.05", "01:22.70"]]}
\end{promptbox}

\subsection{Pseudo-label Generation Prompt}
\label{sec:prompt_pseudo}

GPT-5 generates three (message, cut sequence) pairs per source video
using the prompt below, with frame-grid images as visual context.

\begin{promptbox}{Pseudo-label Generation Prompt}
You are a video editor creating edited videos with clear intent and purpose.

Task:
Create 3 different edited video proposals. For each proposal:

1. First, decide on a specific MESSAGE you want to convey to viewers
   - What story do you want to tell?
   - What emotion or understanding should viewers gain?
   - What is the core intent behind this edit?

2. Then, with that MESSAGE in mind, select the clip intervals (timestamps) that will effectively communicate it
   - Choose clips strategically to build a coherent narrative
   - Each edited video should be 1-3 minutes in total length
   - Use timestamps in [start, end] format with MM:SS:CS (e.g., ['01:23:00', '02:45:50'])
   - You can arrange clips in any order -- chronological order is NOT required

Output exactly 3 EditedVideo objects, each with:
- A SHORT and CONCISE message (5-10 words maximum) that clearly captures the core intent/purpose
- Timestamps selected specifically to deliver that message

IMPORTANT: Timestamp Format (MM:SS:CS):
  MM = minutes (00-59)
  SS = seconds (00-59)
  CS = centiseconds (00-99)
  e.g., "01:30:50" = 1 min 30.50 sec (NOT 1 hour 30 min)
\end{promptbox}

\subsection{MEDitor Prompt}
\label{sec:prompt_meditor}

Our GRPO-trained MEDitor uses the prompt below, kept identical at training and inference time, where \texttt{\{message\}} is the editing message.
Unlike the JSON schema used for the zero-shot baselines, it requests a plain seconds list and states an explicit length budget of roughly 15\% of the video.

\begin{promptbox}{MEDitor Prompt}
You are a professional video editor. Select highlight segments conveying the message below. The selected segments should total roughly 15

Message: "{message}"

List the selected highlights as time ranges in seconds, using exactly this format, with a single period only at the very end:
<start> to <end> seconds; <start> to <end> seconds; ...; <start> to <end> seconds.

Rules:
- Use seconds with one decimal, e.g. "12.5 to 30.0 seconds".
- Separate segments with "; "; put a period only after the final segment.
- Every time must be within the video length and start < end.
- Output only the list and stop right after the final period.

Example:
15.5 to 19.4 seconds; 76.0 to 82.7 seconds.
\end{promptbox}

\subsection{LLM-as-Judge Prompt}
\label{sec:prompt_judge}

The pairwise preference judge (Gemini-3.5-Pro) receives the two
videos followed by the prompt below.
The judge is queried twice per pair (human-first and AI-first) and
the responses are combined to cancel position bias.

\begin{promptbox}{LLM-as-Judge Prompt}
You are an assistant specialized in evaluating the quality of video editing, particularly in terms of story-telling to convey the message.
Evaluate which video's editing more effectively communicates the message: "{message}"

### Criteria:
- Editing Strategy: How the choice of shots, their sequence, and the pacing create a visual narrative for the message.
- Storytelling & Narrative Flow: How well the sequence of shots creates a logical and compelling arc to support the message.
- Informativeness & Visual Detail: The richness of the visual information and how effectively composition, movement, and subjects provide context.

### Output Format (JSON only):
{
  "answer": "A or B",
  "reason": "Explain the effectiveness of the editing and visual storytelling without using labels."
}
\end{promptbox}

\section{Dataset Examples}
\label{sec:dataset_examples}

\begin{figure*}[t]
  \centering
  \includegraphics[width=0.7\textwidth]{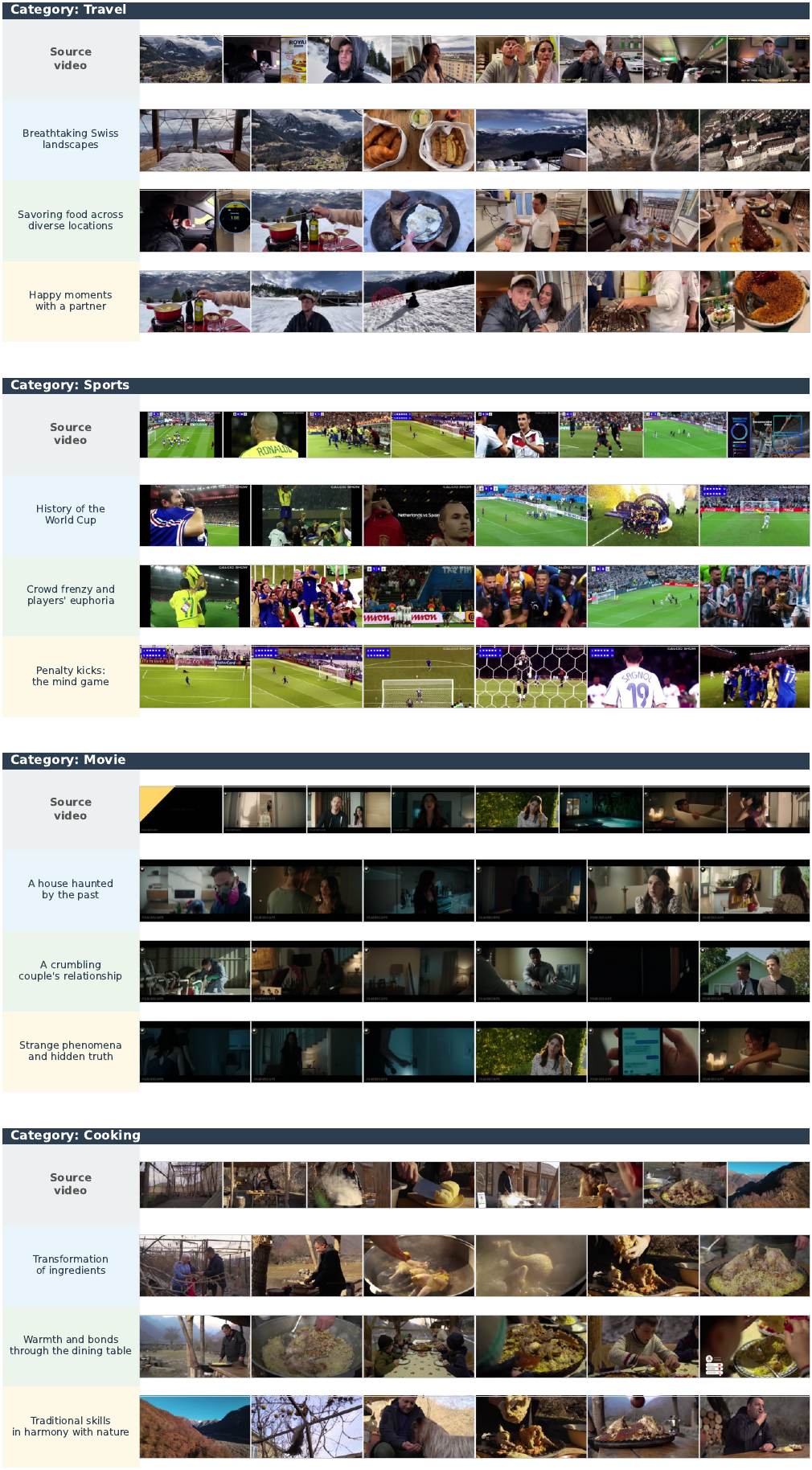}
  \caption{%
    Representative MEDit-Bench examples across four categories (Travel, Sports,
    Movie, Cooking).
    For each source video, the top strip shows 8 uniformly sampled frames
    spanning the full duration ($\sim$14--16\,min).
    The three strips below show frames from a professional editor's cut
    under each of the three editing messages.
  }
  \label{fig:dataset_examples}
\end{figure*}

\Cref{fig:dataset_examples} shows representative frame strips from four
MEDit-Bench source videos alongside their editing messages and corresponding
human-produced cut sequences.
Despite sharing the same source footage, different messages yield
temporally and visually distinct edits, demonstrating the intent-driven
nature of the dataset.
A subset of the edited videos is available on our project page (\url{https://ogatakatsuya.github.io/medit-bench}).

